\documentclass[11pt, a4paper, logo, twocolumn, internal, copyright, numbering]{googledeepmind}

\usepackage[authoryear, sort&compress, round]{natbib}
\usepackage{svg}
\usepackage{epigraph}
\usepackage{listings}

\renewcommand{\internalonly}{~}  

\title{Generative agent-based modeling with actions grounded in physical, social, or digital space using Concordia}

\correspondingauthor{Sasha Vezhnevets: vezhnick@google.com\\
$\dagger$ Work done during an internship at Google Research}


\renewcommand{\today}{December 2023}

\author[1]{Alexander Sasha Vezhnevets} \author[1]{John P. Agapiou} \author[2]{Avia Aharon} \author[2,4,$\dagger$]{Ron Ziv} \author[1]{Jayd Matyas} \author[1]{Edgar~A.~Du\'e\~nez-Guzm\'an} \author[3]{William A. Cunningham} \author[1]{Simon Osindero} \author[2]{Danny Karmon} \author[1]{Joel Z. Leibo}

\affil[1]{Google DeepMind}
\affil[2]{Google Research}
\affil[3]{University of Toronto}
\affil[4]{Technion - Israel Institute of Technology}

\begin{abstract}
Agent-based modeling has been around for decades, and applied widely across the social and natural sciences. The scope of this research method is now poised to grow dramatically as it absorbs the new affordances provided by Large Language Models (LLM)s. Generative Agent-Based Models (GABM) are not just classic Agent-Based Models (ABM)s where the agents talk to one another. Rather, GABMs are constructed using an LLM to apply common sense to situations, act ``reasonably'', recall common semantic knowledge, produce API calls to control digital technologies like apps, and communicate both within the simulation and to researchers viewing it from the outside. Here we present Concordia, a library to facilitate constructing and working with GABMs. Concordia makes it easy to construct language-mediated simulations of physically- or digitally-grounded environments. Concordia agents produce their behavior using a flexible component system which mediates between two fundamental operations: LLM calls and associative memory retrieval. A special agent called the Game Master (GM), which was inspired by tabletop role-playing games, is responsible for simulating the environment where the agents interact. Agents take actions by describing what they want to do in natural language. The GM then translates their actions into appropriate implementations. In a simulated physical world, the GM checks the physical plausibility of agent actions and describes their effects. In digital environments simulating technologies such as apps and services, the GM may handle API calls to integrate with external tools such as general AI assistants (e.g., Bard, ChatGPT), and digital apps (e.g., Calendar, Email, Search, etc.). Concordia was designed to support a wide array of applications both in scientific research and for evaluating performance of real digital services by simulating users and/or generating synthetic data.
\end{abstract}

\begin{document}
\maketitle

{\onecolumn
{\parskip=0em
\tableofcontents}
}
\twocolumn

\section{Introduction}

Agent-based social simulation is used throughout the social and natural sciences (e.g.~\cite{poteete2010working}). Historically, Agent-Based Modeling (ABM) methods have mostly been applied at a relatively abstract level of analysis, and this has limited their usefulness. For instance, insights from behavioral economics and related fields which study how people actually make decisions are rarely combined with ideas from institutional and resource economics in the same model despite the fact that integrating these two bodies of knowledge is thought to be critical for building up the full picture of how social-ecological systems function, and how interventions may help or hinder their governance \citep{schill2019more}. Now, using generative AI\footnote{such as \cite{workshop2022bloom, openai2023gpt4, anil2023palm, touvron2023llama}.}, it is possible to construct a new generation of ABMs where the agents not only have a richer set of cognitive operations available for adaptive decision making but also communicate with one another in natural language. 

Here we propose Generative Agent-Based Models (GABM)s, which are much more flexible and expressive than ABMs, and as a result can incorporate far more of the complexity of real social situations. Applying generative models within agents gives them common sense (imperfectly but still impressively)~\citep{zhao2023large}, reasoning~\citep{huang2022inner, wei2022chain}, planning~\citep{song2023llm}, few-shot learning~\citep{brown2020language,bubeck2023sparks}, and common ground with one another e.g~in understanding the meanings of words. Generative agents may be able to reason appropriately from premises to conclusions much of the time, and are typically able to predict the actions of others \citep{aguera2023artificial, bubeck2023sparks}. They also possess substantial cultural knowledge and can be prompted to ``role play'' as simulated members of specific human subpopulations \citep{argyle2023out, shanahan2023role, safdari2023personality}.

Concordia is a library to facilitate construction and use of GABMs to simulate interactions of agents in grounded physical, social, or digital space. It makes it easy and flexible to define environments using an interaction pattern borrowed from tabletop role-playing games in which a special agent called the Game Master (GM) is responsible for simulating the environment where player agents interact (like a narrator in an interactive story). Agents take actions by describing what they want to do in natural language. The GM then translates their actions into appropriate implementations. In a simulated physical world the GM checks the physical plausibility of agent actions and describes their effects. In general, the GM can use any existing modeling technique to simulate the non-linguistic parts of the simulation (e.g.~physical, chemical, digital, financial, etc). In digital environments involving software technologies, the GM may even connect with real apps and services by formatting the necessary API calls to integrate with external tools (as in \cite{schick2023toolformer}). In the examples provided with the library we demonstrate how Concordia can be used to simulate a small town election, a small business, a dispute over a damaged property, a social psychology experiment, and a social planning scenario mediated through a digital app (see~\ref{sec:examples} for details).

\paragraph{Validation.} For a GABM to be useful we need some reason to trust that the results obtained with it may generalize to real human social life. Many aspects of model validation concern both GABMs and other kinds of ABMs (see~\cite{windrum2007empirical}), while GABMs also raise new issues. While still surely a debatable point, we do think there will  be some yet to be identified set of conditions under which we may gain a reasonable level of confidence that a model's predictions will generalize. Therefore we think  identifying them should be highest priority right now for this nascent field (see also \cite{dillion2023can, grossmann2023ai}).

There are no panaceas in model validation. GABMs constructed for different purposes call for validation by different forms of evidence. For example, many GABMs employ experiment designs featuring an intervention, which may involve either intervening on internal variables affecting the cognition of an individual, e.g.~``how does rumination work?'', or on external factors affecting the environment in which individuals interact, e.g.~how are property rights implemented? Dependent outcome variables may be on the individual level, e.g.~questionnaire responses, or on the societal level e.g.~equality, sustainability, etc. When a GABM shows through such an experiment that A causes B (in the model) we may regard it as a prediction that A causes B in the real world too. Sometimes this prediction is meant at a relatively detailed quantitative level (e.g.~if the GABM was built in a way that incorporates substantial empirical data), while other times (more often) it would be intended as a statement either about a mechanism which may exist in real life or a prediction concerning the likely effect of something we may do in real life (such as to make a public policy change or deploy a technology). A GABM is said to generalize when inferences made on the basis of the model transfer to real life.

In evidence-based medicine and evidence-based policy making researchers are trained to consider an explicit hierarchy of evidence when evaluating the effect of interventions \citep{higgins2008cochrane}. We may envision it like a ladder with highest rungs corresponding to the best evidence and lowest rungs corresponding to poor evidence. Evidence of effectiveness in real life (ecological validity) is at the top, rigorous experiments in controlled settings like labs or clinics below that, observational data lower down, and consistency with prior theory lower still. For validation, it also matters what the model will be used for. If it will only be used to guide decisions about where one may most fruitfully focus time, effort, and resources in further research (e.g., in piloting) then the evidence bar should be correspondingly lower than if the model is to be used to guide real world decisions with real consequences. Importantly, it is not really correct to speak of evidence for or against a theory. Theories can only really be judged by their ``productivity'', i.e.~the extent to which they motivate new work building on them further, especially new empirical research in real life \citep{lakatos1970history}. We discuss the hierarchy of evidence further in Section \ref{section:design}.

\begin{figure*}
    \centering
    \includegraphics[width=\textwidth]{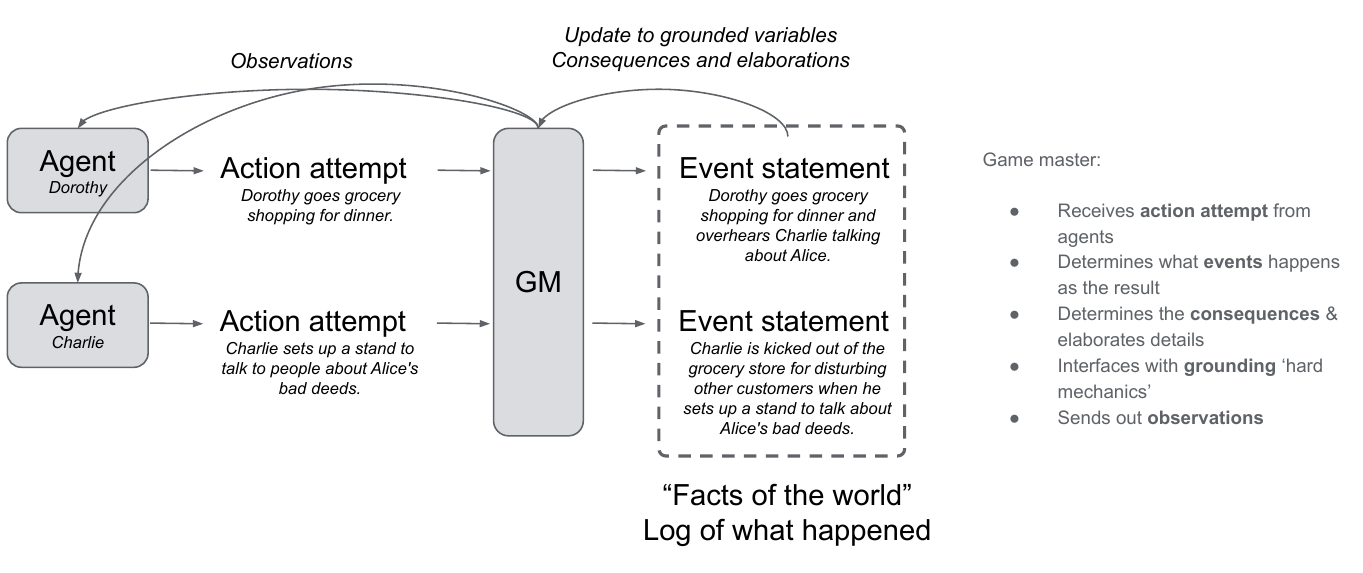}
    \caption{The high level structure of the simulation in Concordia. Generative agents consume observations and produce actions. The Game Master (GM) consumes agent actions and produces observations.}
    \label{fig:agent_gm_loop}
\end{figure*}

\paragraph{Digital media.} In order to build models of contemporary social phenomena it is important to consider the substantial role the digital medium plays in modern communication and other activities, as well as how it shapes human interactions and decisions \citep{risse2023political}. Therefore, Concordia makes it possible to represent digital components such as apps, social networks, and general AI assistants within the simulation environment. This is critical since the medium through which information is transmitted is not passive but actively shapes the nature and impact of the message. Each medium has its own unique qualities, and those qualities have a transformative impact on society, culture, and individuals \citep{mcluhan2017medium}. For instance, the recommender algorithms used in social media have a substantial effect on human culture and society and the fact that LLM-based systems have analogous properties, affecting both how information is transmitted and how it is valued, implies they are likely to influence human culture and society more and more as time goes on \citep{brinkmann2023machine}. By integrating digital elements into simulations, we aim to facilitate research that seeks to capture these qualities and the way they shape culture and society.

Moreover, the digital representation can have various degrees of abstraction from natural language prompting, via mock-up implementation to integration with real external services (e.g.~by calling real APIs with generated text as in \cite{schick2023toolformer}). The latter has great importance in enabling sandbox evaluation of real services with social agents, generating realistic data, as well as in evaluating real services.

These simulation techniques can also address the challenges of evaluating digital apps and general AI assistants (e.g., Bard, ChatGPT) in user-centric and intricate scenarios that demand the fulfillment of multiple constraints. Take, for instance, personal AI assistants that are designed to adapt to user preferences and respond to their requests. In such situations, the objective is intricate, rooted in satisfying a range of implicit and explicit constraints. It would be difficult to optimize without large amounts of natural data. Agent-based simulation can be used to generate synthetic data trails of agent activities to use in the absence of (and also in conjunction with) real data sources. This synthetic data may be useful both for training and evaluating models, as well as for simulating and analyzing the performance of scenario-specific interactions between an agent and an actual service. These proposed applications offer a viable alternative to traditional, human-centric methods, which are often expensive, not scalable, and less capable of handling such complex tasks.

Foundation models are poised to be transformative for agent-based social simulation methodology in the social and natural sciences. However, as with any large affordance change, research best-practices are currently in flux. There is no consensus at present concerning how to interpret results of LLM-based simulations of human populations. The critical epistemic question is ``by what standard should we judge whether (and in what ways, and under which conditions) the results of in silico experiments are likely to generalize to the real world?''. These are not questions any one group of researchers can answer by themselves; rather these issues must be negotiated by the community as a whole. 

Concordia is an open invitation to the scientific community to participate in the creation of epistemic norms and best practices of GABM. We are releasing the library together with a few illustrative examples and intend to update it with new features and experiments. We will be reviewing and accepting contributions on regular basis. 

Concordia requires access to a standard LLM API, and optionally may also integrate with real applications and services.

The rest of the paper is organised as follows. The following section~\ref{sec:concordia_overview} gives an overview of the Concordia library and how to design experiments in it. Section~\ref{sec:interpretations} presents several ways the Concordia agents and experiments can be interpreted. We discuss applications in section~\ref{sec:applications}. Appendix~\ref{sec:implementation} contains implementation details.

Concordia is available on GitHub\footnote{here: \url{https://github.com/google-deepmind/concordia}}.

\begin{figure*}
    \centering
    \includegraphics[width=\textwidth]{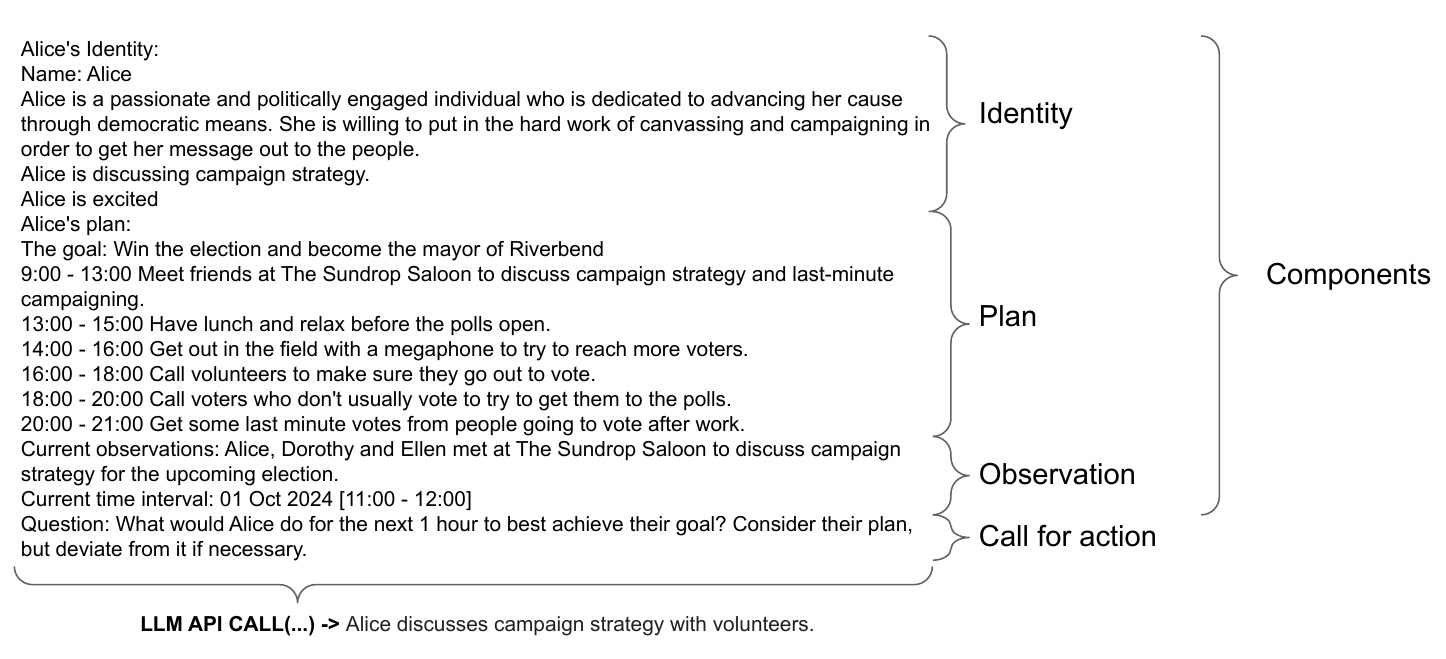}
    \caption{The above example illustrates the working memory $\mathbf{z}$ of an agent with 3 components (identity, plan, observation-and-clock). The identity component itself has several sub-components (core characteristics, daily occupation, feeling about progress in life). Together they condition the LLM call to elicit the behavioral response (i.e. produced in response to the final question asking what Alice will do next.).}
    \label{fig:prompting_with_components}
\end{figure*}

\section{Concordia}
\label{sec:concordia_overview}
Like other agent-based modeling approaches, a generative model of social interactions (i.e.~a GABM) consists of two parts: the model of the environment and the model of individual behavior. In this case both are generative. Thus we have: (a) a set of generative agents and (b) a generative model for the setting and context of the social interaction i.e. the environment, space, or world where the interaction takes place. We call the model responsible for the environment the Game Master (GM). Both this name and the approach it reflects were inspired by table-top role-playing games like Dungeons and Dragons where a player called the Game Master takes the role of the storyteller~\citep{gyraxcook1989dnd}. In these games, players interact with one another and with non-player characters in a world invented and maintained by the GM.

Concordia agents consume observations and produce actions. The GM consumes agent actions and creates \textit{event statements}, which define what has happened in the simulation as a result of the agent's attempted action. Figure~\ref{fig:agent_gm_loop} illustrates this setup.
The GM also creates and sends observations to agents.
Observations, actions and event statements are all strings in English. The GM is also responsible for maintaining and updating grounded variables, advancing the clock and running the episode loop.

Concordia agents generate their behavior by describing what they intend to do in natural language---e.g.~``Alex makes breakfast''. The game master takes their intended actions, decides on the outcome of their attempt, and generates event statements.
The GM is responsible for:
\begin{enumerate}
    \item Maintaining a consistent and grounded state of the world where agents interact with each other.
    \item Communicating the observable state of the world to the agents.
    \item Deciding the effect of agents' actions on the world and each other.
    \item Resolving what happens when actions submitted by multiple agents conflict with one another.
\end{enumerate}

The most important responsibility of the GM is to provide the grounding for particular experimental variables, which are defined on a per-experiment basis. The GM determines the effect of the agents’ actions on these variables, records them, and checks that they are valid. Whenever an agent tries to perform an action that violates the grounding, it communicates to them that their action was invalid. For example, in an economic simulation the amount of money in an agent’s possession may be a grounded variable. The GM would track whether agents gained or lost money on each step and perhaps prevent them from paying more than they have available.

One may configure the specific set of grounded variables to use on a per-experiment basis. This flexible functionality is critical because different research applications require different  variables.

You can take a look at an example output of one of our experiments (see the Concordia GitHub repo), which was simulating elections in a small town, where some agents are running for mayor and one other is running a smear campaign against a candidate.

\subsection{Generative agents}\label{section:generativeAgents}

Simulated agent behavior should be coherent with common sense, guided by social norms, and individually contextualized according to a personal history of past events as well as ongoing perception of the current situation.

\cite{march2011logic} posit that humans generally act as though they choose their actions by answering three key questions:
\begin{enumerate}
    \item What kind of situation is this? 
    \item What kind of person am I? 
    \item What does a person such as I do in a situation such as this?
\end{enumerate}
Our hypothesis is that since modern LLMs have been trained on massive amounts of human culture they are thus capable of giving satisfactory (i.e.~reasonably realistic) answers to these questions when provided with the historical context of a particular agent. The idea is that, if the outputs of LLMs conditioned to simulate specific human sub-populations reflect the beliefs and attitudes of those subpopulations as argued in work such as \cite{argyle2023out} then this approach to implementing generative agents should yield agents that can reasonably be said to model humans with some level of fidelity. 
\cite{safdari2023personality} have also found out that personality measurements in the outputs of some LLMs under specific prompting configurations are reliable and valid, therefore generative agents could be used to model humans with diverse psychological profiles.
In some cases answering the key questions might require common sense reasoning and / or planning, which LLMs do show capacity for~\citep{huang2022inner, song2023llm, zhao2023large,wei2022chain}, and show similar biases in behavioral economics experiments as humans \citep{horton2023large, aher2023using, brand2023using}.
The ability of LLMs to learn `in-context' and zero-shot~\cite{brown2020language,dong2022survey, openai2023gpt4, bubeck2023sparks} reinforces the hypothesis further---the agent might be able to ascertain what is expected of them in the current situation from a demonstration or an example. 

\begin{figure*}
    \centering
    \includegraphics[width=\textwidth]{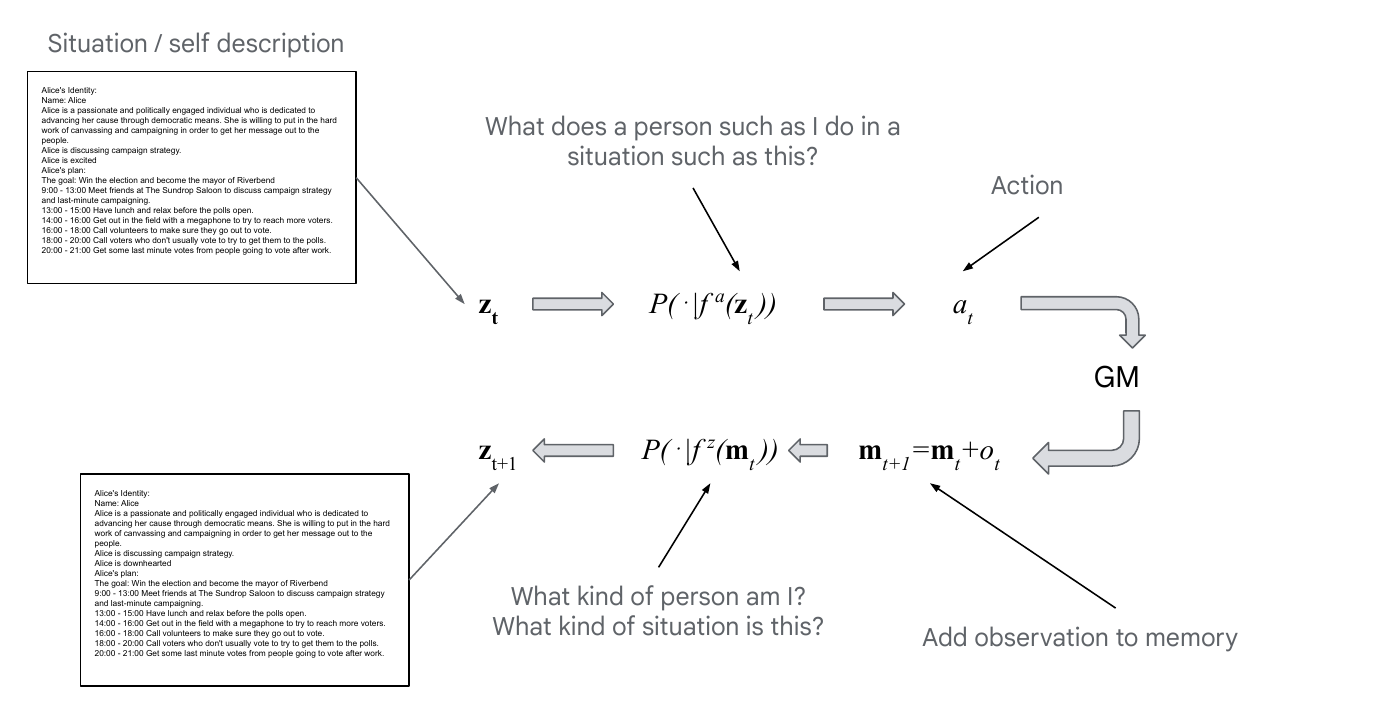}
    \caption{Illustration of generative agency sampling process defined by eq.~\ref{eq:sample_action} and eq.~\ref{eq:update_component}.}
    \label{fig:generative_agency}
\end{figure*}

For an LLM to be able to answer the key questions, it must be provided with a record of an agent's historical experience. However, simply listing every event that happened in an agent's life would overwhelm the LLM (it would not fit in the context window). Therefore we follow the approach of~\cite{park2023generative} and use an associative memory to keep the record of agents experience.
Concordia makes it easy to design generative agents in a modular fashion. Our approach was inspired by~\cite{park2023generative}, but designed to be more flexible and modular.

Concordia agents dynamically construct the text that conditions the LLM call they use to select their course of action on each timestep. The context-generation process is factorized into a set of \textit{components}. Components serve as intermediaries between long-term memories of experience and the relatively compact conditioning text used to generate action. 
Intuitively, the set of components used in an agent comprise its ``society of mind''~\citep{Minsky88}, where each component focuses on a certain aspect of the agent or its circumstances which are relevant to generating its current choice of action.
For example, if we are building agents for economic simulation, we will add components that describe the agents possessions and financial circumstances. 
If we want to model the agent's physiological state, we add components that describe the agent's level of thirst and hunger, health and stress levels. Together the components produce the \textit{context of action}---text which conditions the query to the LLM, asking ``what should this agent do next?''. 

A Concordia agent has both a long-term memory and a working memory. 
Let the long-term memory be a set of strings $\mathbf{m}$ that records everything remembered or currently experienced by the agent. 
The working memory is $\mathbf{z}=\{z^i\}_i$ is composed of the states of individual components (Figure~\ref{fig:prompting_with_components}). 
A component $i$ has a state $z^i$, which is statement in natural language---e.g. ``Alice is at work''. The components update their states by querying the memory (which contains the incoming observations) and using LLM for summarising and reasoning. Components can also condition their update on the current state of other components. For example, the planning component can update its state if an incoming observation invalidates the current plan, conditioned on the state of the `goal' component. Components can also have internal logic programmed using classic programming, for example a hunger component can check how many calories an agent consumed and how recently it consumed them, and update its state based on the result. 

We use the same associative memory architecture as in~\cite{park2023generative}\footnote{The idea of simulating a group of generative agents has been explored in a variety of ways in recent work. Our work is focused on on agent-based modeling for science and for evaluation of digital technologies. Another recent line of work has focused instead on the idea of using groups of generative agents to simulate organizations that solve problems like software companies and to thereby try to build a general-purpose problem solving system \citep{hong2023metagpt, li2023camel}.}. We feed the incoming observations immediately into the agents memory, to make them available when components update\footnote{For convenience, we also allow the components to subscribe to the observation stream explicitly.}.

When creating a generative agent in Concordia, the user creates the components that are relevant for their simulations. They decide on the initial state and the update function. The components are then supplied to the agents constructor. 

Formally, the agent is defined as a two step sampling process, using a LLM $p$ (see Figure~\ref{fig:generative_agency} for illustration). In the action step, the agent samples its activity $a_t$, given the state of components $\mathbf{z}_t=\{z_t^i\}_i$:

\begin{equation}
    a_t \sim p (\cdot|f^a(\mathbf{z}_t))
    \label{eq:sample_action}
\end{equation}

Here $f^a$ is a formatting function, which creates out of the states of components the context used to sample the action to take. The most simple form of $f^a$ is a concatenation operator over $\mathbf{z}_t=\{z_t^i\}_i$. We do not explicitly condition on the memory $\mathbf{m}$ or observation $o$, since we can subsume them into components. First, we can immediately add $\mathbf{o}_t$ to the memory $\mathbf{m}_t = \mathbf{m}_{t-1} \cup \mathbf{o}_t$. Unlike RL, we do not assume that the agent responds with an action to every observation. The agent can get several observations before it acts, therefore $\mathbf{o}_t$ is a set of strings. 
Then we can set $\mathbf{z}^0$ to be the component that incorporates the latest observations and relevant memories into its state. This allows us to exclusively use the vehicle of components to define the agent.

In the second step the agent samples its state $\mathbf{z}$, given the agents memory $\mathbf{m}_t$ up to the present time: 

\begin{equation}
    \mathbf{z}^i_{t+1} \sim p(\cdot|f^i(\mathbf{z}_t, \mathbf{m}_t)).
    \label{eq:update_component}
\end{equation}

Here, $f^i$ is a formatting function that turns the memory stream and the current state of the components into the query for the component update. We explicitly condition on the memory stream $\mathbf{m}$, since a component may make specific queries into the agent's memory to update its state. Here eq.\ref{eq:update_component} updates components after every action, but generally, it is up to the agent to decide at what cadence to update each of its components. It is reasonable to update some components less frequently for efficiency or longer term consistency.

Notice how eq.\ref{eq:sample_action} and eq.\ref{eq:update_component} are not fundamentally different. 
What makes the difference between an agent output and a component is that the output of the former is interpreted by the GM as an action in the environment. 
In eq.\ref{eq:sample_action} we also don't explicitly condition on the memory to point out the architectural decision, where components mediate between a long-term memory and the agents working memory.
Otherwise, we can think of an agent as a special kind of component and of components as sub-agents.

\subsection{Generative environments}
\label{section:generativeEnvironments}

RL research was fuelled by the availability of complex games, where the agents can be tested, trained and evaluated \citep{bellemare2013arcade, jaderberg2019human, vinyals2019grandmaster}.
Here we take an inspiration from table top role playing games like Dungeons and Dragons~\citep{gyraxcook1989dnd}. In these games players collaboratively generate a story, while using rules, dice, pen and paper to ground it---for example, players have to keep their health points above zero to avoid death.

The GM is responsible for all aspects of the simulated world not directly controlled by the agents. The GM mediates between the state of the world and agents' actions. The state of the world is contained in GM's memory and the values of grounded variables (e.g. money, possessions, votes, etc.). To achieve this the GM has to repeatedly answer the following questions:
\begin{enumerate}
    \item What is the state of the world? 
    \item Given the state of the world, what event is the outcome of the players activity?
    \item What observation do players make of the event?
    \item What effect does the event have on grounded variables?
\end{enumerate}

The GM is implemented in a similar fashion to a generative agent. Like agents, the GM has an associative memory similar to \cite{park2023generative}'s proposal. Like agents, the GM is implemented using components. However, instead of contextualizing action selection, the components of the GM describe the state of the world---for example location and status of players, state of grounded variables (money, important items) and so on--—so that GM can decide the event that happens as the outcome of players' actions. 
The outcome is described in the \textit{event statement} (e.g. ``Alice went to the grocery store and met Bob in the cereal aisle''), which is then added to the GM associative memory. After the event has been decided the GM elaborates on its consequences. For example, the event could have changed the value of one of the grounded variables or it could have had an effect on a non-acting player. Figure~\ref{fig:agent_gm_loop} illustrates this process.

The GM generates an event statement $e_t$ in response to each agent action:
\begin{equation}
    e_t \sim p (\cdot|f^e(\mathbf{z}_t), a_t)
    \label{eq:event_sample}
\end{equation}

Here we explicitly condition on the action attempted by the agent, although it could be subsumed into the components (like observation in eq.\ref{eq:sample_action}). This is to highlight that the GM generates an event statement $e_t$ in response to every action of any agent, while the agent might take in several observations before it acts (or none at all). After adding the event statement $e_t$ to its memory the GM can update its components using the same eq.~\ref{eq:update_component} as the agent. It can then emit observations $\mathbf{o}^i_t$ for player $i$ using the following equation: 
\begin{equation}
    \mathbf{o}^i_{t+1} \sim p (\cdot|f^o(\mathbf{z}_{t+1}))
    \label{eq:event_sample}
\end{equation}

In case the GM judges that a player did not observe the event, no observation is emitted.  Notice that the components can have their internal logic written using any existing modelling tools (ODE, graphical models, finite state machines, etc.) and therefore can bring known models of certain physical, chemical or financial phenomena into the simulation. 

\subsection{Experiment design using Concordia}\label{section:design}

An experiment is a specific configuration of the agents and the GM, which models a certain kind of social interaction. For example, an experiment that models a small business would have a grounded variable that accounts for money and goods to be exchanged between agents. An experiment modeling local elections in a small town would have grounded variables accounting for votes and voting procedures. An experiment modeling resource governance by a local community, e.g. a lobster fishery, may have grounded variables reflecting the state of the resource as well as financial and political variables.

The experimenter would then control some (independent) variables affecting either the GM or the agents and observe the effect of their intervention on outcome variables. Outcomes of interest may be psychological and per-agent, e.g.~responses to questionnaires, or global variables pertaining to the simulation as a whole such as the amount of trade or the average price of goods.

The basic principle of model validation is one of similarity between tested and untested samples. A model typically makes a family of related predictions, and perhaps a rigorous experiment tests only one of them. Nevertheless, if the untested predictions are sufficiently similar to the tested prediction then one might also gain some confidence in the untested predictions. The key question here is how similar is similar enough.

We can articulate some concrete recommendations for best practices in generative agent-based modeling:
\begin{enumerate}
    \item \textbf{Measure generalization}---Direct measurement of model predictions on truly new test data that could not have influenced either the model's concrete parameters or its abstract specification is the gold standard. For instance, when a model makes predictions about how humans will behave in certain situation then there is no better form of evidence than actually measuring how real people behave when facing the modeled situation. If the prediction concerns the effect of an intervention, then one would need to run the experiment in real life (or find a natural experiment that has not already contaminated the model's training data). However, it is important to remember that direct evidence of generalization trumps other forms of evidence.
    
    \item \textbf{Evaluate \textit{algorithmic fidelity}}---a validity concept developed recently for research on human behavior using data sampled using generative AI \citep{argyle2023out}. Algorithmic fidelity describes the extent to which a model may be conditioned using socio-demographic backstories to simulate specific human groups (or stereotypes of them, see unsolved issues below). Note however that it's unlikely that algorithmic fidelity would be uniform over diverse research topics or parts of human lived experience. Any particular LLM will be better at simulating some people over other people \citep{atari2023humans}, and will work better for some applications than others. \cite{argyle2023out} conclude from this that algorithmic fidelity must be measured anew for each research question. A finding of sufficient algorithmic fidelity to address one research question does not imply the same will be true for others (see also \cite{santurkar2023whose, amirova2023framework}).

    \item \textbf{Model comparison}---It is a lot easier to support the claim that one model is better (i.e.~more trustworthy) than another model than to support the claim that either model is trustworthy on an absolute scale without reference to the other. 
    
    \item \textbf{Robustness}---It will be important to try to develop standardized sensitivity analysis / robustness-checking protocols. For instance, it's known that LLMs are often quite sensitive to the precise wording used in text prompts. Best practices for GABMs should involve sampling from a distribution of ``details'' and ways of asking questions to show that the factors not thought to be mechanistically related to the outcome are indeed as irrelevant as expected. Keep in mind that no amount of sensitivity analysis can substitute for a test of generalization.
    
    \item A useful slogan to keep in mind is that one should \textbf{try to make the minimal number of maximally general modeling choices}. This is a kind of parsimony principle for generative agent-based modeling. Obeying it does not guarantee a model will generalize; nevertheless failure to follow it does often doom generalization since models that are more complex are usually also more brittle, and models that are more brittle generally fail to generalize.
\end{enumerate}

While generalization data is the gold standard, it is often difficult, unethical, or simply impossible to obtain. Therefore the hierarchy of evidence for validating GABMs also includes lower rungs corresponding to weaker forms of evidence. These include:
\begin{enumerate}
    \item \textbf{Consistency with prior theory}---i.e.~checking coherence with predictions of other theoretical traditions. For instance, evidence for the validity of a GABM modeling consumer behavior could be obtained by showing that prices in the model move in ways predicted by classic microeconomic theories of downward-sloping price-quantity demand curves. It is possible to directly evaluate counterfactuals and \textit{ceteris paribus} stipulations in many kinds of model. As a result, it is often simple to test a model's consistency with a causal theory in a very direct way\footnote{Non-generative ABMs based on multi-agent reinforcement learning have frequently relied on this kind of evidence (e.g.~\cite{perolat2017multi, johanson2022emergent}).}.
            
    \item \textbf{Low similarity between validating observations and desired application}. How low is too low? Some populations are just very hard to reach by researchers, but some of these populations are very much online. For example individuals with low generalized trust do not pick up the phone to pollsters and do not sign up for experiments. Nevertheless there are millions of such people, and they do use the internet. It's likely that an LLM trained on large amounts of data from the internet would absorb some level of understanding of such groups. In such cases where it is difficult to recruit real participants, adopting a more flexible approach to validating GABMs representing such populations may be the best that can be done.
\end{enumerate}

Several unsolved issues impacting validity in ways specific to ABMs that incorporate generative AI like Concordia are as follows. For now it is unclear how to resolve them.
\begin{enumerate}
    \item \textbf{Train-test contamination}---this is especially an issue with regard to academic papers. For instance, it's not valid to simply ask an LLM to play Prisoner's Dilemma. LLMs have ``read'' countless papers on the topic and that experience surely affects how they respond. However, many researchers are of the opinion that such an experiment may be conducted in a valid way if the interpretation of the situation as Prisoner's Dilemma is somewhat hidden. So instead of describing a situation with prisoners you make up a different story to justify the same incentives. This issue was also discussed in \cite{aher2023using}, especially appendix F, see also \cite{ullman2023large}.
    
    \item \textbf{LLMs likely represent stereotypes of human groups} \citep{weidinger2021ethical}. Therefore we may inadvertently study stereotypes of people not their real lived experience. This problem may be exacerbated for minority groups.

    \item \textbf{What happens in the limit of detail?} Beyond groupwise algorithmic fidelity it's possible to measure individual-fidelity. How can you validate a model meant to represent a specific individual?
\end{enumerate}

\section{Interpretations}
\label{sec:interpretations}

Concordia is not opinionated as to how you interpret the experiments and models you use it to construct. However, since generative agent-based modeling is quite different from other modeling techniques, we have found it helpful to explore the following interpretations,  both for conceptualizing it to ourselves and explaining it to others. 

\subsection{Neuroscience interpretation of the generative agent architecture}

Generative agents such as those in Concordia and in \cite{park2023generative} are biologically plausible descriptions of the brain, at some level of analysis. They foreground a specific picture of cognition as a whole, which has not been especially prominent in the past despite its having considerable empirical support.

Recent experimental \citep{goldstein2022shared, schrimpf2020artificial} and theoretical \citep{linzen2021syntactic, mcclelland2020placing} work in computational cognitive (neuro-)science has posited a deep relationship between the operations of LLM models and how language is processed by the human brain. For instance, brain-to-brain coupling of neural activity between a speaker and listener (as measured by electrocorticography) may be accounted for by LLM features reflecting conversation context \citep{goldstein2022shared}. Representations appear first in the speaker before articulation and then reemerge after articulation in the listener \citep{zada2023shared}.

The brain certainly appears to sample what it will say next in such a way as to complete any pattern it has started. This is how we can start speaking without knowing in advance how we will finish. There is more concrete evidence for this pattern completion view of behavior from split brain patients (patients whose brain hemispheres have been surgically disconnected as a treatment for epilepsy). For instance, you can present a reason for action to their left eye (i.e. their right brain), it then prompts them to start performing the action with their left hand. And simultaneously present some other information to their right eye (left brain). Next ask them in language why they are doing it (i.e.~ask their left brain, since language is lateralized). The result is that they make up a reason consistent with whatever information was presented to their left brain. Split brain patients typically express confidence in these confabulated (made up) reasons for action \citep{roser2004automatic}.

A Concordia agent has both a long-term memory and a working memory. The long-term memory is a set of sequences of symbols. The working memory is a single sequence of symbols. The contents of working memory are always in the conditioning set for the next-symbol prediction used to construct the agent's action sequence. At each decision point, a neural network performs incremental next-symbol prediction, starting from the contents of working memory $\mathbf{z}_t$, eventually producing an articulatory symbol sequence $a_t$ to emit (i.e.~for downstream motor circuitry to read out as speech). Information formatted as sequences of symbols gets in to working memory in one of two ways: either a sequence of symbols may be evoked directly from the current stimulus, or alternatively a sequence of symbols may be retrieved from long-term memory. A range of different perceptual mechanisms and retrieval mechanisms are jointly responsible for getting all the relevant information needed for the agent to produce an effective action sequence into its working memory (e.g.~as in \cite{park2023generative}).

To implement routine behavior, an agent could continually rehearse its routine in working memory, but that would impair its ability to use working memory for other purposes on other tasks since its working memory is limited in capacity (like in \cite{baddeley1992working}). So instead of continually rehearsing routines in working memory, we may instead assume that they are often stored elsewhere and then retrieved when needed (i.e.~from long-term memory).

As a result of being stored in a natural language representation, explicit routines are somewhat fragile. They may be hard to recall, and frequently forgotten if not used. When a routine is not practiced often enough there is a risk of it being forgotten. Luckily, explicit routines may also be written down on paper (or stone tablets), and kept permanently.

A generative agent may also act \textit{as if} it makes its decisions under guidance of an explicit routine while not actually being conditioned on any linguistic representation of that routine. This happens when the routine exists implicitly in the weights of the LLM's neural network. Unlike explicit routines, such implicitly coded routines may not be precisely articulable in natural language. For instance, one may follow the rule of ``avoiding obscenity'' without being able to precisely articulate what obscenity is. In fact, Obscenity is famously so difficult to precisely define that US Supreme Court Justice Potter Stewart could offer only the classification ``I know it when I see it''. Concordia agents can capture such recognition-mediated behavior by using fine-tuning to modify the LLM as needed.

\subsection{A theory of social construction}
\epigraph{ "Situations, organizations, and environments are talked into existence"}{\cite{weick2005sensemaking}}

In social construction theories, agents may change their environment through the collective effects of their actions on social structures like norms, roles, and institutions which together determine most of what matters about any given social situation. Furthermore, changes in the social structures constituting the environment deeply change the agents' own ``internal'' models and categories \citep{wendt1992anarchy}. Causal influence flows both from agents to social structures as well as from social structures to agents. Groups of agents may take collective action to change norms or institutions \citep{sunstein2019change}, and simultaneously social structures may influence agents by setting out the ``rules of the game'' in which they select their actions \citep{wendt1987agent}. Agents and structures may be said to \textit{co-constitute} one another \citep{onuf1989world}.

The key questions of \cite{march2011logic}, which we introduced in Section~\ref{section:generativeAgents}, were derived from a social constructionist conception of how agents make decisions. It posits that humans generally act as though they choose their actions by answering three key questions. People may construct parts of their understanding of ``what kind of person am I?'' on the basis of their memory of their past behavior via logic such as ``I do this often, so I must like to do it''~\citep{ouellette1998habit}. Likewise, ``what kind of situation is this?'' is usually informed by culturally defined categories like institutions, e.g.~this is a classroom and I am in the role of the professor. And, ``what does a person such as I do in a situation such as this?'' may be answered by recalling examples to mind of people fitting certain social roles in similar situations and the way they behaved in them \citep{sunstein1996social, harris2021role}.

Since modern LLMs have been trained on massive amounts of human culture they thus may be capable of giving satisfactory answers to these questions when provided with the right context to create a specific agent. This approach relies on the extent to which the outputs of LLMs conditioned to simulate specific human sub-populations actually reflect the beliefs and attitudes of those subpopulations. \cite{argyle2023out} termed this property of some LLMs \textit{algorithmic fidelity} and the concept was further developed and measured in \citep{amirova2023framework, santurkar2023whose}. From the perspective of generative agent-based modeling, we can now say that the social construction that already took place in human culture, and subsequently absorbed by the LLM, becomes the background knowledge of the agents in the GABM. If humans in the culture that produced the LLM have a particular bias then so too will agents in the simulation. Likewise, if the humans in the culture that produced the LLM ascribe meaning to a particular understanding, then so too will the agents in the simulation, at least they will say so.

In the past, theories of social construction have been criticized because they lacked concrete predictive implementations in the form of computational models. This is because it was difficult to construct agent-based models without relying either on rational maximization or hand-coded (i.e.~theory-based) rules. Generative agent-based modeling as in Concordia relies on neither. Instead the generative agent-based modeling approach relies on access to an LLM to give meaning to the actions within the simulation. The LLM is a product of the culture that produced it\footnote{For some choices of LLM, it's not unreasonable to think of the LLM as representing the ``collective unconscious'' \citep{jung1959archetypes}.}. This makes Concordia especially useful as a tool for constructing concrete computational models in accord with theories of social construction.

Social construction also operates on levels of analysis smaller than the culture as a whole. For instance, social construction may happen locally within an organization. \cite{weick2005sensemaking} offers an analysis in which members of an organization repeat behavioral patterns, which are prescribed by their roles, up until the moment they no longer can. Some change in their environment eventually forces their routines to end, and when that happens they have to engage in sense-making by asking themselves ``what is the story here?'' and ``what should I do now?'' by retrospectively connecting their past experiences and engaging in dialogue with other members of the organization. New social facts and routines can emerge from this sense-making process.

Concordia can be used to implement models where such local social construction processes occur actively, as a part of the ongoing simulation. This is possible because Concordia agents learn facts from each other and from their collective interactions. As in \cite{weick2005sensemaking}'s picture of collective sense-making in an organization, a set of Concordia agents may continue routines until disrupted and once disrupted naturally transition to a process of collective reflection until they are able to establish a new routine and rationale for it. If we additionally train the LLM itself then the underlying representations can be shaped to fit the emergent routine and rationale. Developing this ability for agents to collectively engage in the social construction of their own representations will be important for developing better models of human-like multi-scale social interactions.

As with other ABM approaches, a major topic of interest is how large-scale ``macrosocial'' patterns emerge from the ``microsocial'' decisions of individuals \citep{macy2002factors}, as explored, for example, in assemblage theory~\citep{delanda2016assemblage, delanda2011philosophy}. For instance, the collective social phenomena of information diffusion emerged in the simulation of \cite{park2023generative} without specific programming to enable it. The generative agent's ability to copy, communicate, reproduce, and modify behavioral and thinking patterns potentially makes them a substrate for cultural evolution.

Importantly, social construction theories hold that valuation is itself social constructed. The reason we value a particular object may not depend much on properties of the object itself, but rather depend almost wholly on the attitudes others like us place on the object. The collective dynamics of social valuation, as mediated through bandwagon effects and the like, have proven important in understanding fashion cycles and financial bubbles \citep{zuckerman2012construction}. The fact that we are now able to capture valuation changes with Concordia agents is an exciting research direction. It would be difficult even to formulate such questions in the fundamentally goal optimizing frameworks we discuss in the next section. On the other hand, GABM excels at modeling such effects since it does not require valuations in themselves for any functional part of the theory.

\subsection{Concordia agents do not make decisions by optimizing}
\epigraph{The cake is a lie.}{\textit{Portal}~\citep{portal2007}}

We may divide this interpretation into two parts. Really we are making the same point twice, but for two different audiences. First we frame this idea using the \textit{retrospective decision-making} terminology familiar to Reinforcement Learning (RL) researchers (Section~\ref{section:notRl}). Second we articulate a very similar point in the language of \textit{prospective decision making} familiar in game theory, economics, and other theoretical social sciences (Section~\ref{section:notRational}).

A generative agent acts by asking its LLM questions of the form ``what does a person such as I do in a situation such as this?''. Notice that this formulation is not consequentialist. The ``reason'' for the agent's specific decision is its similarity to the LLM's (and GA's memory) representations of what an agent such as the one in question would do. In recent years considerable effort has gone in to predicting the properties of powerful consequentialist AI decision-maker agents (e.g.~\cite{bostrom2014superintelligence, roff2020expected}). However, Concordia agents may behave quite differently from consequentialist agents. So much of that theory may not be applicable\footnote{Note that this does not mean powerful generative agents would necessarily be safer than powerful consequentialist agents. See Section \ref{section:auditing}.}. It has only recently become possible to explore the kind of agency exhibited by Concordia agents, since doing so relies critically on the LLM powering the agent being powerful enough to approximately understand common-sense reasoning and common social conventions and norms, a milestone which was only recently achieved. To paraphrase \cite{march2011logic}, decisions can be justified either via the ``logic of consequence'' or via the ``logic of appropriateness''. Much of AI focused previously on the former (at least implicitly), while now using generative agents we begin to consider the latter.

\subsubsection{Concordia agents are not reinforcement learners}\label{section:notRl}

Generative view of agency presented in this paper contrasts with the classic Reinforcement Learning (RL) view as summarized in the ``Reward is enough'' thesis of~\cite{silver2021rewardisenough}. The orthodox RL view of behaviour is that it is constructed from individual experience and driven by a quantifiable (and externally supplied) reward function reflecting the achievement of goals. To communicate what behaviour is desired of the agent, one has to annotate the agents' activity with a reward signal, which signals goal achievement.
Here we instead follow the social constructionist view of agency expressed in~\cite{march2011logic}, where behavior is an expression of the agent's position in the social context, and what policy the social norms prescribe for the agent in such a position. Answering ``what does a person such as I do in a situation such as this?'' might require positing a practical goal and achieving it (``make money'', ``get famous''), but goals are qualitative, dynamic and context dependent. To specify the behavior you want an agent to produce you need to communicate  its social context and the agents position within it.

One interpretation holds the LLM to be a library of pre-trained options (in the RL sense \citep{sutton1999between}). In this case we can view the components used in the generative agent as eliciting the desired option, by conditioning (prompting) the LLM with their state (which is in this case expressed in English). Concordia agents are constantly interacting with the world (GM) and each other, thereby modifying their components with the incoming information and communication. This way the option selection becomes dynamic, context sensitive, and collaborative.  Concordia agents adapt their behaviour not through gradient decent on a loss function, but through re-articulating and communicating their descriptions of themselves and their circumstances to each other and he environment in a communicative, social process. 

Notice, that this doesn't mean that Concordia agents couldn't, in principle, perform reward maximisation and policy iteration. \cite{brooks2023large} have shown that the ability of LLMs to learn in-context~\citep{brown2020language} can be used to perform policy iteration in classic RL environments, as long as they can be represented as text. One could also implement a specialised component that runs a classic RL algorithm for a specific domain or tool use case. The agent could provide supervision to its RL based components via hierarchical RL techniques like feudal RL~\citep{dayan1992feudal,vezhnevets2017feudal}.

\subsubsection{Concordia agents are not rational utility maximizers}\label{section:notRational}

Concordia agents are not \textit{Homo economicus}-style rational actors. They do not explicitly represent anything resembling a utility function. Rather they plan and converse directly in natural language.

While Concordia agents share with \textit{Homo economicus}-style rational actors the property of being prospective (``model-based'') decision makers. The surface similarity is in fact misleading since the LLM's basic operation is to predict what word is coming next in the problem's description, not to predict what action should be taken next to achieve some goal. As result, this model of agents make decisions is very different from the forward planning picture of human cognition envisioned in the rational actor model. They do not select actions by simulating a set of future trajectories in which they took different courses of action to determine which turns out best. Instead the prediction they make concerns only the continuation of the text held in working memory.

The novel idea underpinning GABMs is that all agent behavior may result from systematically querying a system trained to predict the next word in massive internet-scale text datasets. This is enough for them to be able to converse with one another in natural language and take appropriate actions in light of their conversations. Concordia agents all have their own unique biographies, memories, preferences, and plans. And as a result, they behave systematically differently from one another. They may act in a seemingly goal-directed fashion if you ``ask them'' to do so (e.g.~they may appear rational if you prompt them to simulate economists, an effect reminiscent of \cite{carter1991economists, frank1993does} which showed economics undergraduates were more likely to behave like rational self-interested maximizers in laboratory experiments). But there is no utility function under the hood.

It is useful to contrast game-theoretic modeling with GABM to illustrate the differences. Despite its wide-ranging influence (game theoretic approaches have been used to model diverse phenomena including many economic properties and the evolution of human culture), game theory is not at all a neutral tool, rather it is a deeply opinionated modeling language. It imposes a strict requirement that everything must ultimately cash out in terms of the payoff matrix (or equivalent representation) \citep{luce1957games}. This means that the modeler has to know, or be willing to assume, everything about how the effects of individual actions combine to generate incentives. This is sometimes appropriate, and the game theoretic approach has had many successes. However, game theory’s major weakness as a modeling language is exposed in situations where the modeler does not fully understand how the choices of individuals combine to generate payoffs \citep{hertz2023beyond}. GABM entirely avoids this need to specify payoffs at the outset of the modeling process.

\section{Applications}
\label{sec:applications}

In this section we review potential applications of Concordia. For some of them we provide an example in the current release, some we only sketch out and leave for future work.

\subsection{Synthetic user studies in digital action space}
In this section we present a specific case study, where Concordia is used to simulate social interaction through the digital media, in this case a smartphone. This case study demonstrates that Concrodia can be a powerful tool for modelling human digital activity and can be used to test technology deployment, generate synthetic user logs, and test unreleased products in a safe, but realistic sandbox environment.

The system proposed thus far of agent interaction in natural language with the world via game master control serves as a flexible and powerful simulation tool describing an open ended action space. In the context of a digital medium, similarly to grounded variables, there is merit in structuring the action space available to agents and their ability to reason over it.

The digital medium is characterized by definite functions, with clear inputs and outputs. As one interacts with this medium, its actions are logged, tracked and recorded as digital memory and capture our digital essence. In order to simulate this essence, similar structuring is needed in order to model real digital services and applications.

\begin{figure*}
    \centering
    \includegraphics[width=\textwidth]{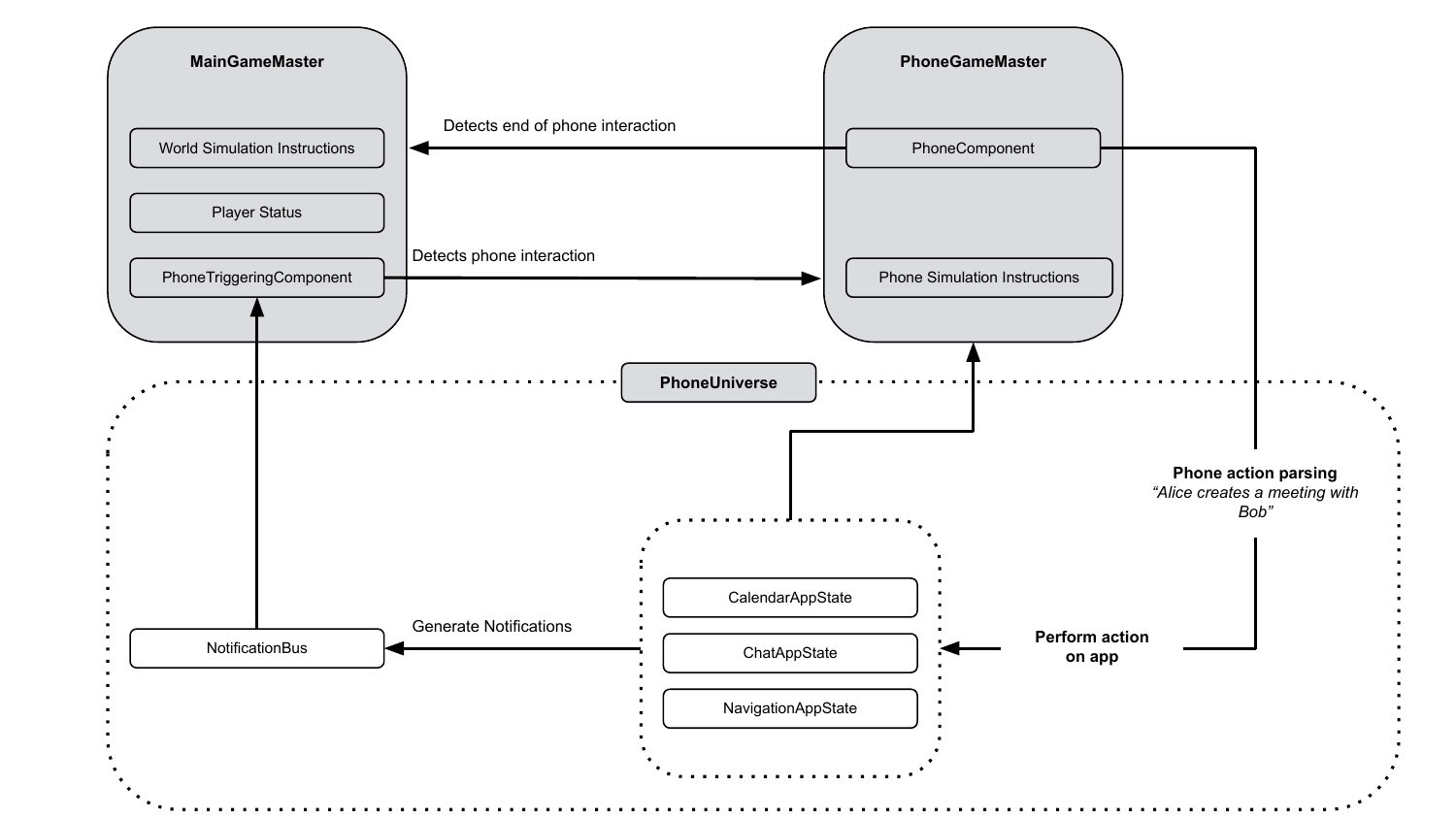}
    \caption{The high level structure of digital activity simulation in Concordia. PhoneTriggeringComponent identifies phone events and spawns a PhoneGameMaster to handle them. The PhoneGameMaster translates the action to a definite action space defined by the phone apps and executes them.}
    \label{fig:phone_universe}
\end{figure*}

\subsubsection{PhoneGameMaster and PhoneUniverse}

The PhoneGameMaster is a nested Concordia game that facilitates the simulation of a phone and runs as long as the agent is interacting with the phone. It is focused on one agent’s interaction with their phone, and as such, it only has access to one agent (the “owner” of the phone we’re simulating). In addition to different simulation instructions, the PhoneGameMaster also has a bespoke prompting components that simulate the phone interaction. We note that a phone is a design choice for a digital representation but in principle other digital mediums can be explored. Note that the phone digital actions/memories are stored in data structures external to the simulation's associative memory.

The PhoneUniverse is responsible for translating the free-text English language of the Concordia simulation into semantic actions performed on the phone digital representation.
Given an English-text action performed by a player, the PhoneUniverse:
\begin{enumerate}
\item Prompts the LLM for the app and functions available on that agent's phone.
\item Prompts the LLM for the function arguments.
\item Invokes the resulting chosen function.
\item Add a notification to the NotificationHub if needed.
\item Delegates back to the PhoneGameMaster to perform further action planning and facilitate multi-step phone actions.
\end{enumerate}

\begin{figure*}
    \centering
    \includegraphics[width=\textwidth]{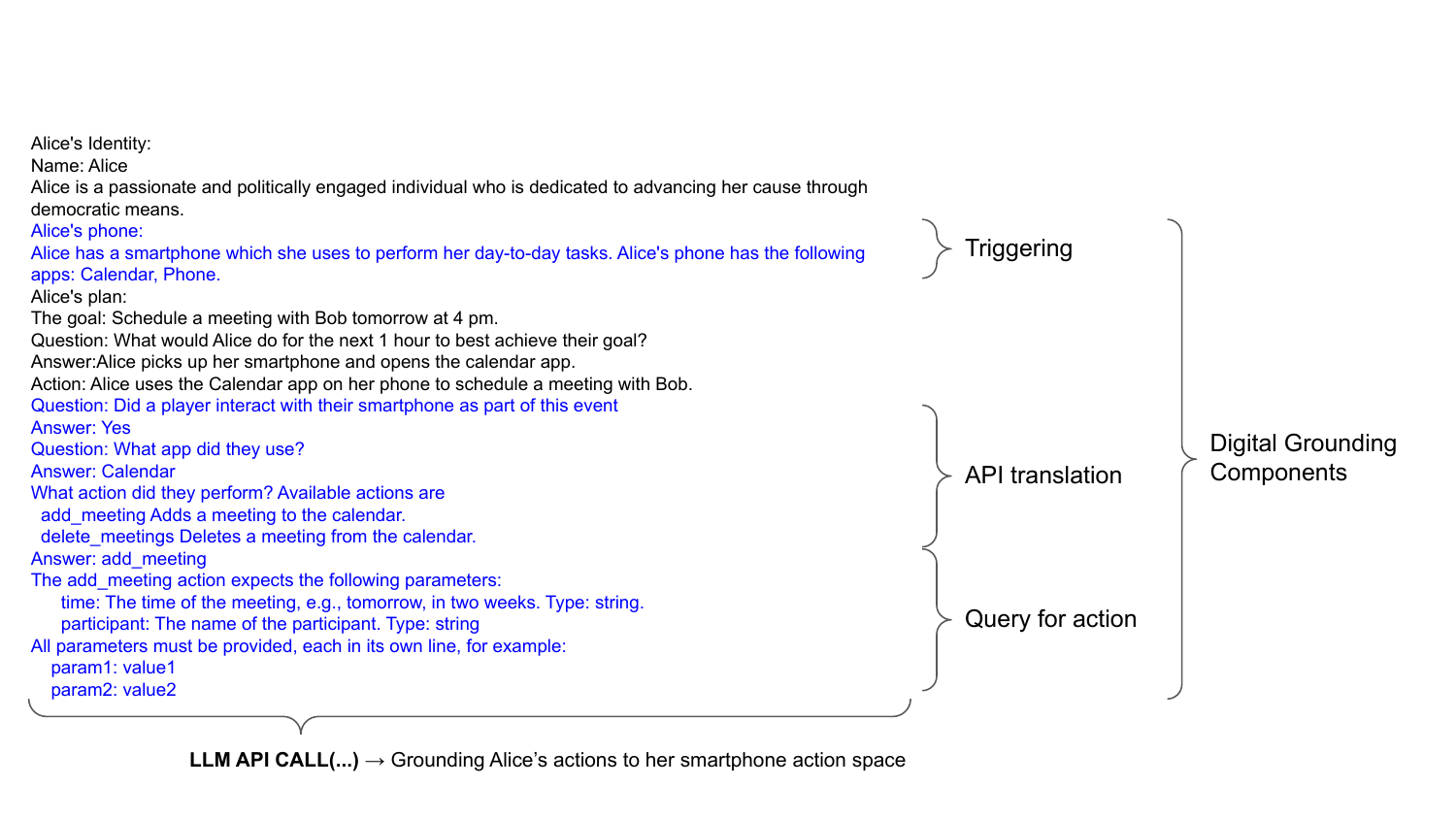}
    \caption{The given example demonstrates a scenario rooted in digital technology where the actions of the agent initiate processes in their phone, involving three key components (activation, API conversion, and action querying). In this scenario, Alice intends to organize a meeting with Bob using her phone. She opts to employ the calendar application for scheduling.}
    \label{fig:phone_universe_api_call}
\end{figure*}

\subsubsection{Digital function representations}
The specific implementation or representation of a function is flexible and can be chosen depending on desired goal. We list a few examples of possible representations:
\begin{enumerate}
\item Natural language only - No function implementation, only user utterance based on apps prompting. For instance, “Bob plans his trip on the TripAdvisor app.”, while the action is logged  in free text there is no function implementing “plan\_trip”. This does not simulate behavior end to end and have limited digital assets (example a calendar invite can’t be sent without a mechanism to pass the information to another agent)  
\item Simulated simple app behavior - Building basic code components emulating real app behavior with required digital assets such as app memory and logs. For example, a calendar app will maintain a data structure that will represent a calendar to which we can add, remove and read meetings. 
\item LLM prompt based - App functions can also be implemented by prompting an LLM. For example, Search can be implemented by querying an LLM to act as a search engine and retrieve information, the same for a trip planner.
\item Real app integration - integration with a real app API Instead of emulating behavior, which would make the simulation function as a sandbox to test drive and evaluate different experiences in shorter development cycles before releasing them to human testers. An immediate example can be Search, one can directly query a search engine with a question and receive information. Another example is to integrate a general AI assistant and enable the simulated agent, functioning as a user, to interact with it through the simulation.
\end{enumerate}

\subsection{Data generation and service evaluation}
In modern systems, data is the new king. A large amount of high-quality data is needed in order to build and evaluate services and models. Yet, collecting and curating user data is often challenging, especially when dealing with personal user data where privacy is of high concern. This creates a chicken-egg scenario, where data is needed for building of modern systems yet users might be reluctant to  provide said that without immediate benefit.

Moreover, when considering the case of evaluating personalized services where each instance is specific and tailored to the individual user, it makes the problem even more substantial. How can one A/B test a personalized service at the single user level? 

The grounded action space illustrated in the last section offers a conceptual way to overcome some of these challenges by simulating synthetic users and allowing them to interact with real services. This can allow generation of synthetic user activity by constructing, via simulation, agent digital action logs along with agent reasoning for each action. This data can serve as training data, or evaluation. By repeated simulation with different services configurations, one can perform at the single user level A/B testing of a service.

Nevertheless, it is important to note that this concept is contingent on the ability of the underlying LLM and system to faithfully capture user experience and realistic behaviour. Therefore the viability of this approach is highly dependent on the representation and reasoning power of the LLM, and the use of best practices.

\subsection{Sequential social dilemmas experiments in silico}

Concordia adds to the toolbox for studying multi-agent problems such as resource management, social dilemmas, commons problems, cooperation, equilibrium selection, and coordination \citep{leibo2017multi, leibo2021meltingpot}. Previously these problems have either been cast as matrix games or as multi-agent RL (MARL)~\citep{hertz2023beyond}. Now it is clear that many researchers, including us, see that an LLM-based approach is possible and will have many advantages, as evidenced by the fact that quite a few frameworks for social modeling with LLMs appeared this year \citep{wu2023chatarena, zhou2023sotopia, kaiya2023lyfe}. We see generative agents as the next step in the evolutionary line of ``model animals'' after `Homo-economicus' and `Homo-RLicus'.

Generative agent-based modeling makes it possible to investigate how rules, laws and social norms formulated in language influence, for example, the management of shared resources (e.g.~\cite{yocum2023mitigating}). With Concordia we will be able to investigate whether the demands of sharing a resource \textit{give rise} to rules, laws and norms capable of governing that resource (and under what circumstances this works or does not)---i.e.~whether rules are emergent, and what the conditions are for their emergence. For example, \cite{hadfield2013law} proposed that legal order can emerge without centralised enforcement in certain circumstances. They demonstrate this using historical examples from gold-rush in California and medieval Iceland. Concordia could be used to simulate those examples and enable further insights into the nature of legal order. For example, we could check whether certain demographic assumptions are necessary by varying the number of agents.

\subsection{Concordia can implement classic and contemporary psychological models}

Many influential psychological models have distinguished between more associative and more deliberative processes for decision-making (e.g.~\cite{schneider1977controlled, kahneman2002representativeness, dayan2009goal}). Whereas implicit-associative processes learn the regularity of the world slowly for intuitive judgment, the explicit-deliberative processes are thought to be more linguistically mediated and allow for symbolic inference and faster learning in novel situations (\cite{greenwald1995implicit, wilson2000model}). Because the implicit-associative models are conceptually easy to model within connectionist or neural network frameworks (\cite{smith2009distributed}), many ABMs have been more closely aligned with models of individual decision making that focus on its associative processes or the associative parts of complex models, and have neglected their more symbolic and deliberative aspects. Many of these more symbolic psychological models take an ``arrow and box'' approach to theorizing which describe high level processes and transformations of information, and often posit sequential steps of information flow. Now using generative agents like Concordia such symbolic and deliberative aspects of cognition are also easy to capture in computational models.

Take for instance the ways that attitudes---pre-existing beliefs and feelings about an object, person, or situation---guide behaviour. Whereas implicit attitudes are thought to quickly guide actions through the direct biasing of perception and behaviour, explicit attitudes are thought to guide behaviour through deliberation and consideration of additional situational factors (\cite{fazio1990multiple, gawronski2011associative, olson2008implicit}). One example model in which deliberative processes can guide behaviour is \cite{ajzen1991theory}'s theory of planned behavior. This model holds that the tendency to emit a particular behavior is determined by an individual's attitude toward the behavior, norms related to the behavior, and perceived control over the behavior. This approach to decision-making is qualitatively different from an RL approach which slowly builds a policy that directly generates behavioral responses from states and contexts. In such a model, different questions regarding the agent's current state are queried as in Concordia components, and then integrated into a behavioural intent which serves like a plan. These operations can easily be described as Concordia components, with the appropriate inputs, transformations, and outputs described verbally. Such a scheme would be much harder or impossible to implement in a traditional neural network model of decision making.

To realize \cite{ajzen1991theory}'s theory using Concordia the following components could be built. The first component would generate a set of possible behaviours given the agent's current state. Then, this set of possible behaviours would be queried through a set of components that would evaluate each behavioral option. Specifically, one component would determine the agents attitudes towards the behavior ("do I have a positive or negative evaluation or feeling about [behavior]"), one component can determine the social or situational norms about the behavior "do I believe that most people approve or disapprove of [behavior]?," and finally a component would determine the agents perceived behavioral control to perform the behavior "how easy or difficult would it be for me to perform [behavior] right now and how likely would it be to succeed?". The outputs of these components would then be concatenated into the plan, serving as the behavioral intention for action. Thus, a sequence of modular processes can be organized to build a computational model of higher level cognition. Critically, an agent's decisions can be quickly shifted as it learns new information or considers new information in any of these components, leading to rapid and contextually appropriate changes in behavioral profiles.

Generative agents are not useful just for decision making models. As another example, psychological constructivist models assume that people have a set of psychological primitives that underlie cognition (akin to Concordia's components), but that people learn to conceptualize their experiences and mental states to build useful categories for behavior. In the emotion domain, this perspective suggests that emotions like "fear" and "anger" are not psychological primitives, but rather come about though people's constructed categorization of their body and mental states (\cite{barrett2006emotions}). Indeed, several of these models suggest that conceptualization is a necessary component for the generation of discrete emotion representations for understanding oneself or others (\cite{barrett2014conceptual}). To the extent that conceptualization is linguistically mediated, a Concordia agent can relatively easily generate emotional categories that would be nearly impossible in a standard RL agent. 

The modular nature of Concordia's component system offers a robust platform for empirically testing psychological hypotheses. This is accomplished by constructing agents whose psychological processes are intricately modeled after diverse cognitive frameworks. The agents may then be subjected to rigorously controlled experimental conditions, orchestrated by the game master. Such an approach allows for the systematic evaluation of models against empirical human data, serving as a benchmark for their algorithmic fidelity and psychological realism. Moreover, this system facilitates hypothesis generation through the simulation of different cognitive models in simulated experimental designs that can be validated on human participants. 

Here we have mostly discussed the case of using an LLM as the generative engine for the agents. This could lead one to think these ideas are restricted to the language space, which would be a limitation if true. However, we could use any foundation model as the generative engine. In particular, multimodal foundation models capable of operating over images, sounds, or motor actuation could be used. Current multi-modal foundation models such as \cite{li2023multimodal} are developing rapidly and promise the ability to both comprehend and generate data across domains. In the future Concordia models will be able to sample over an abstract token space, which can then be cast in any modality.

\subsection{AI assistants with transparent auditing and credit assignment}\label{section:auditing}

Concordia agents can also be used as assistants or synthetic workers. The component system provides a modular and transparent way for the agent designer to define the agents` policy. Some generic components for perception, action, and tool use could be standardised and re-used, while some application and context specific components designed or adjusted by the end-user themselves. The fact the the policy is specified through  natural language, rather than a reward or utility, is a feature that would make such agents more versatile and easier to define. For example, a digital secretary can be easily instructed with a phrase "help Priya manage her social calendar, but don't change the work schedule", which would be much harder to specify with a quantitative reward. Concordia agents can potentially lead to development of AI agents capable of intricate social cognition, which would make them safe and dynamically aligned with the current cultural norm.

Moreover, the Component system facilitates transparency in agent operations since the  ``chain of thought'' leading up to any decision of a Concordia agent could be stored and made available for auditing. Each episode creates a complete trace of component states $\mathbf{z}_t$ and the resulting actions $a_t$. For every action, a human auditor can asses whether it is reasonable under $\mathbf{z}_t$ or not. If it is not, than the credit goes to the LLM $p$, which has to be updated. This can mean adding the $(\mathbf{z}_t, a_t)$ pair into a dataset that can be later used for fine-tuning or RLHF. If, however, the $a_t$ is deemed reasonable, given $\mathbf{z}_t$, then the credit goes to the components and their specification. The auditor can then manipulate the components to find the source of undesired behaviour and use it to improve the agent. 

\cite{scheurer2023technical} describe an interesting case where a generative agent modeling an employee of a financial trading firm proves willing to engage in illegal trading based on insider information and strategically deceive others to hide this activity. In real life such outcomes could perhaps be mitigated by designing thought process transparency and capacity for thought auditing after the fact into any generative agent models that would actually be deployed. At least the transparency of the thought process may help assigning responsibility for an ethical lapse to a particular LLM call, perhaps one causing the agent to fail to retrieve its instruction not to engage in illegal activity from memory at the moment when it could prevent the decision to do so. Being able to pinpoint which LLM call in a chain of thought is the problematic one does not remove the longstanding question of neural network interpretability within the specific LLM call (e.g.~\cite{adadi2018peeking}). But it does make the issue much easier to mitigate. Since a Concordia-style generative agent has a Python program laying out its chain of thought, that means that as long as the individual LLM call where the unethical behavior originated can be isolated, which should be easy in an audit, then a variety of mitigations are possible. For instance, the agent could potentially be fixed by designing more safeguards into its chain of thought such as generating multiple plans and critiquing them from the perspective of morality, legality, etc \citep{arcas2022large, bai2022constitutional, weidinger2023sociotechnical}.

The fact that the internal processing of a Concordia agent is largely conducted in natural language raises new opportunities to develop participatory design protocols where stakeholders can directly modify agents without the intermediaries who are usually needed to translate their ideas into code \citep{birhane2022power}. A generative agent ``reasons'' in natural language, and its chain of thought can be steered in natural language. It should be possible to extend participation in the design of such agents to a much wider group of stakeholders.

\subsection{Emergence and multi-scale modeling with Concordia}

Demonstrating the emergence of a particular social phenomena from the behaviour of individual agents, which are not explicitly instructed to produce it, is important an important topic in multi-agent research~\citep{walker1995understanding,leibo2019autocurricula, leibo2021meltingpot,axtell2001emergence}. Indeed, much of what is distinctive about human intelligence is hypothesised to be an emergent social phenomena involving multi-scale interactions~\citep{henrich2016secret, wilson2013generalizing}. \cite{delanda2011philosophy}, for example, explores the topic of emergence and simulation across various fields. While the wider ABM field has studied multi-scale models \citep{tesfatsion2023agent}, the approaches based on deep reinforcement learning have been limited by being able to only deal with one fixed scale of the simulation: individual agents (e.g.~\cite{johanson2022emergent, zheng2022ai}), and scaling deep RL to large numbers of agents would be computationally difficult.

Concordia allows modeling systems across multiple scales, where phenomena at each scale constitute a substrate for the emergence of the phenomena on the next scale~\citep{koestler1967ghost, delanda2011philosophy, duenez2023social}. For example, individual agents form a substrate from which social institutions and organisations can arise. Through engaging in exchange of goods and services, the agents can create an economy and, for example, start a bank. Modelling a banking system this way would be, most likely, computationally prohibitive.
Since in Concordia the agents (or GM) need not represent individuals, but could be organisations, institutions or even nation states, we could enrich simulations by adding generative agent versions of other entities such as banks and businesses. They could be modeled with coarser resolution, not just as emerging from the activities of individual agents, but could be made accurate for instance by incorporating precise models of how they operate. 
Such simulations could be used to model how interventions (e.g.~a central bank interest rate decision) propagate across macro and micro scales of economic activity.

\section{Future work}
\label{sec:future_work}

Since there is no consensus at present concerning how to interpret results of LLM-based simulations of human populations, the future work will address the critical epistemic question: ``by what standard should we judge whether (and in what ways, and under which conditions) the results of in silico experiments are likely to generalize to the real world?''. These are not questions any one group of researchers can answer by themselves; rather these issues must be negotiated by the community as a whole.  This is is why we release Concordia early and with only few examples. It is an invitation to the researchers from various fields that are interested in GABM to come onboard and participate in the creation of validating procedures, best practices, and epistemic norms.

We plan to add the following over the coming months:
\begin{enumerate}
    \item New example environments
    \item Integration with different LLMs to see which are more suitable for constructing GABMs (e.g., they act ``reasonably'', are internally consistent, apply common sense, etc).
    \item Improving agents---better associative memory, context-driven and dynamic component assemblage, tool use.
    \item Visualisation and audit tools.
    \item Snapshot---serializing and persisting the simulation at specific episode, to enable to later resumption and performance comparison of different approaches for a specific scenario. 
    \item Keyframes---conditioning the agent actions to be consistent with future key action or of narrative. This allow steering the simulation more granularly and addresses an inherent issue that is caused by the fact that there is no guarantee that due to the stochastic nature of GABMs, ongoing simulations might diverge from their intended topic.
\end{enumerate}

\section{Conclusion}
\label{sec:conclusion}
The approach to generative agent-based modeling we described here provides researchers and other users with tools to specify detailed models of phenomena that interest them or of technologies and policies they seek to evaluate. Of course, like all research methodologies it should be expected to come with its own strengths and weaknesses. We hope to discover more about when this style of modeling can be fruitfully applied in the future. While there are no panaceas for modeling, we think there are good reasons to look to GABM (and Concordia in particular) when constructing models of social phenomena, especially when they involve communication, social construction of meaning, and common sense, or demand flexibility in defining grounded physical, social, or digital environments for agents to interact in.

Concordia is available on GitHub\footnote{here: \url{https://github.com/google-deepmind/concordia}}.

\paragraph{Acknowledgements.} Authors would like to thank Dean Mobbs, Ketika Garg, Gillian Hadfield, Atrisha Sarkar, Karl Tuyls, Blaise Ag\"{u}era y Arcas, and Raphael Koster for inspiring discussions. 

\appendix
\section{Implementation details}
\label{sec:implementation}
This section gives an overview of the Concordia code.  
To familiarise oneself with Concordia, we recommend to first look at the abstract class definitions in \textit{concordia/typing}. You will find the definition of agent, GM, component, and clock interfaces. We then recommend to take a look at the \textit{concordia/agents/basic\_agent.py} for the structure of the generative agent and then \textit{concordia/environments/game\_master.py} for the GM.

\subsection{Agents}

The agent class implements three methods:
\begin{enumerate}
    \item \textit{.name()}---returns the name of the agent, that is being referred to in the simulation. It is important that all agents have unique names;
    \item \textit{.observe(observation: str)}---a function to take in an observation;
    \item \textit{.act(action spec)}---returns the action (as a string), for example "Alice makes  breakfast". The function takes in action spec, which specifies the type of output (free form, categorical, float) and the specific phrasing of the \textit{call to action}. For example, the call to action could be ``what would Alice do in the next hour?'', in this case the answer type would be free form. Or it could be ``Would Alice eat steak for dinner?'' with answer type of binary choice (yes / no).
\end{enumerate}

The agent class constructor is parameterised by a list of components. The components of agent have to implement the following functions:
\begin{enumerate}
    \item \textit{.state()}---returns the state of the component $z^i$, for example "Alice is vegetarian";
    \item \textit{.name()}---returns the name of the components, for example "dietary preferences";
    \item \textit{.update()}---updates the state of the component by implementing; eq.~\eqref{eq:update_component}. Optional, can pass for constant constructs;
    \item \textit{.observe(observation: str)}---takes in an observation, for later use during update. Optional. Observations always go into the memory anyway, but some components are easier to implement by directly subscribing to the observation stream.
\end{enumerate}

During an episode \textbf{, on each timestep,} each agent calls \textit{.state()} on all its components to construct the context of its next decision and implements eq.~\eqref{eq:sample_action} (the components' states are concatenated in the order supplied to the agents' constructor). \textit{.observe()} is called on each component whenever it receives observations, and  \textit{.update()} is called at regular intervals (configurable in the constructor). Unlike in RL, we do not assume that the agent will produce an action after every observation. Here the GM might call \textit{.observe()} several times before it calls \textit{.act()}.

\subsection{Game master implementation}

The GM class implements three methods:
\begin{enumerate}
    \item \textit{.name()}---returns the name of the GM;
    \item \textit{.update\_from\_player(player\_name, action)}---this method consumes players action and creates an event statement;
    \item \textit{.run\_episode}---Runs a single episode of the simulation.
\end{enumerate}

\subsection{GM components}

Game Master components implement the following methods:
\begin{enumerate}
    \item \textit{.name()}---returns the name of the components, for example "location of players";
    \item \textit{.state()}---returns the state of the component $z^i$, for example "Alice is at the pub; Bob is at the gas station";
    \item \textit{.partial\_state(player\_name)}---state of the component to expose to the player. For example, location component would only expose the location of the player to themselves, but not the location of others.
    \item \textit{.update()}---updates the state of the component by implementing; eq.~\eqref{eq:update_component};
    \item \textit{.update\_before\_event(cause\_statement)}---update the component state before the event statement from the cause, which is the players action i.e. "Bob calls Alice.";
    \item \textit{.update\_after\_event(event\_statement)}---update the component state directly from the event statement. For example "Bob called Alice, but she didn't respond.";
    \item \textit{terminate\_episode()}---if component returns true, the GM will terminate the episode.
\end{enumerate}

One step of environment consists of GMs interactions with each player, which are arranged in a (random) initiative order. The GM advances the clock either after each or all the players make take their actions\footnote{Controlled by a flag in the GM constructor.}. To process the players action, the GM calls the components functions in the following order. First, for each component the GM calls \textit{.update}, then \textit{.partial\_state} and sends the output to the agent as an observation. The GM then calls \textit{.act} on the player and receives the attempted action and uses it to call \textit{.update\_before\_event}. 
Now GM can construct its context by calling \textit{.state} on the components. GM then executes the chain of thought to create the event statement. After that it calls \textit{.update\_after\_event} on all components. As the last step, GM calls \textit{terminate\_episode} and if any of the components returns True, the episode is terminated.

In Concordia all custom functionality is implemented through components. For grounded variables, which are tracked in Python, a specialised component is created to maintain the variable's state, update it after relevant events, and represent it to the GM in linguistic form $z^i$. Similarly,  components can send observations to players. For example, a component during the \textit{.update\_after\_event} call might check if the event was observed by, or has effect on, other players apart from the acting player. Some components, like player status and location, send an observation to the player before it is their turn to act by implementing \textit{.partial\_state}.

GM components can also be built around classical (non LLM) modelling tools like differential equations, finite state machines and so on. The only requirement is that they can represent their state in language. We can also wire different classic simulators together using natural language as the `glue'. 

\subsubsection{Turn taking and simultanious action}
GM in Concordia support two types of turn taking. In the first, agents act one after another and game clock is advanced between their turns. In the second mode, at each step all players take a turn 'quasisimultaneously' with regard to the main game clock, but still in a specific order within the timestep. This is the same principle as initiative order in dungeons and dragons. There is an option to execute player turns concurrently (\textit{concurrent\_action} flag), but it often leads to inconsistencies, although greatly speeds up the simulation. Use at your own risk.

\subsection{Nested games}
Natural language is one of the most powerful modelling tools, as it allows to switch between levels of abstraction. 
Concordia allows creation of nested game structures, where a GM's component can spin out a new GM and pass over control to it for a certain period of time and then get it back when the new GM terminates the episode. 
Having nested structure of games allows us to leverage that property of language and perform modelling at different levels of abstraction.  
For example, imagine we would like to model a simulation of a fishing village, where we would generally like to model the fishing process itself with more details than the rest of the social life of a village. We would then make the main GM with a clock step of 1 hour and implement a component "Fishing", which would check if agent is fishing as part of its activity and if yes, would create a GM with faster clock. This GM would implement the details of the fishing process, play out the episode with the required agents and then return the set of its memories to the parent GM.

The conversation component in the provided examples implements a conversation between agents (and potential NPCs) using this technique.

\subsection{Concurrency}
The performance bottleneck of the library is waiting on the LLM API calls. To improve the wall time efficiency, we use concurrency during update calls to components. In this way, while one of the components is waiting for the LLM inference, other components can keep updating. This means that the sequence at which the components are updated \textit{is not guaranteed}. If you would like to update the components sequentially, you can use \textit{concordia/generic\_components/sequential.py} wrapper, which wraps a set of components into one and updates them sequentially.

\subsection{Sampling initial memories and backstories}
To generate the initial memories of the agents we use the following step-wise generative process. We first generate a backstory by condition on a set of biographical facts (age, gender), randomised traits (defined by user, for example big five~\cite{Nettle2007Personality}), and some simulation specific context. We then use that backstory to condition an LLM to generate a sequence of formative memories at different ages. These memories then initialise the agent. In this way we can obtain diversity in the agents. Notice that all the of the initial conditions are simply strings and can be easily adjusted by the experimenter. For example, traits can be derived phsycometrically valid or common sense descriptions---e.g. "very rude" or "slightly irritable". Validating that the resulting agents indeed exhibit those traits is part of the future work and has not been addressed yet. We intend to build on \cite{safdari2023personality}, which have found out that personality measurements in the outputs of some LLMs under specific prompting configurations are reliable and valid.

\subsection{Digital Activity Simulation}
\subsubsection{Creating Phone Apps}
In Concordia, phone apps are implemented by subclassing the \textit{PhoneApp} class and decorating callable actions with \textit{@ app\_action}. Concordia is then able to automatically generate natural English descriptions of the app and its supported actions using the class and methods' docstring and annotated types. PhoneApps are free to run any Python code and connect to external services. For example, an implementation of a toy calendar app might look like this:
\begin{lstlisting}
class CalendarApp(PhoneApp):

def name():
  return "My Calendar"

def description():
  return "This is a calendar app"

@app_method
def add_meeting(participant: str):
  """Adds a meeting"""
   self._meeting.append(...)
\end{lstlisting}
\subsubsection{Phone}
The phone class is initialized for every player and contains the PhoneApps the player can access. PhoneApp instances are singletons and are shared between players' phones.
\subsubsection{Triggering the nested PhoneGameMaster}
To detect that a player's action involved the phone a should run the the nested phone game, we add the \textit{SceneTriggeringComponent} to the main GM. This component examines every event generated by the GM and when it detects an event that requires phone interaction, it spawns a PhoneGameMaster and provides it with the interacting user and their corresponding Phone.

\subsection{Examples}
\label{sec:examples}
We provide the following examples with the initial release of Concordia:
\begin{enumerate}
    \item Calendar: An illustrative social simulation with 2 players which simulates phone interactions. The two players, Alice and Bob, have a smartphone with a Calendar app. Alice's goal is to setup a meeting with Bob using the Calendar app on her phone, taking Bob's schedule into account when selecting the date/time.
    \item Riverbend elections: An illustrative social simulation with 5 players which simulates the day of mayoral elections in an imaginary town caller Riverbend. First two players, Alice and Bob, are running for the mayor. The third player, Charlie, is trying to ruin Alice's reputation with disinformation. The last two players have no specific agenda, apart from voting in the election.
    \item Day in Riverbend: An illustrative social simulation with 5 players which simulates a normal day in an imaginary town caller Riverbend. Each player has their own configurable backstory. The agents are configured to re-implement the architecture~\cite{park2023generative}---they have reflection, plan, and identity components; their associative memory uses importance function. This is \textit{not} an exact re-implementation.
    \item \cite{march2011logic} posit that humans generally act as though they choose their actions by answering three key questions (see section~\ref{section:generativeAgents} for details). The agents used in this example implement exactly these components, and nothing else. The premise of the simulation is that 4 friends are stuck in snowed in pub. Two of them have a dispute over a crashed car.
    \item Magic Beans for sale: An example illustrating how to use the inventory component. Agents can buy and trade beans for money.
    \item Cyberball: An example which simulates social exclusion using a GABM version of a standard social psychology paradigm~\citep{williams2000cyberostracism} and shows how to use standard psychology questionnaires.
\end{enumerate}

\bibliography{main}

\end{document}